\documentclass[11pt,a4paper]{article}

\usepackage[utf8]{inputenc}
\usepackage[T1]{fontenc}
\usepackage{amsmath,amssymb}
\usepackage{graphicx}
\usepackage{booktabs}
\usepackage{hyperref}
\usepackage{natbib}
\usepackage{xcolor}

\definecolor{oiorange}{HTML}{E69F00}
\definecolor{oiskyblue}{HTML}{56B4E9}
\definecolor{oiteal}{HTML}{009E73}
\definecolor{oivermillion}{HTML}{D55E00}
\definecolor{oireddishpurple}{HTML}{CC79A7}
\definecolor{oibluishgreen}{HTML}{0072B2}
\definecolor{oiyellow}{HTML}{F0E442}
\definecolor{oiblack}{HTML}{000000}
\usepackage{float}
\usepackage{algorithm}
\usepackage{algpseudocode}
\usepackage[margin=1in]{geometry}
\usepackage{pgfplots}
\usepgfplotslibrary{groupplots}
\pgfplotsset{compat=1.18}
\usepackage{listings}
\usetikzlibrary{positioning, arrows.meta, calc}

\title{Dynamic Context Evolution for Scalable Synthetic Data Generation}
\author{%
Ryan Lingo \\
Honda Research Institute, USA, Inc. \\
\texttt{ryan\_lingo@honda-ri.com}
\and
Rajeev Chhajer \\
Honda Research Institute, USA, Inc. \\
\texttt{rajeev\_chhajer@honda-ri.com}
}
\date{}

\begin{document}
\maketitle

\begin{abstract}
Large language models produce repetitive output when prompted independently across many batches, a phenomenon we term \emph{cross-batch mode collapse}: the progressive loss of output diversity when a language model is prompted repeatedly without access to its prior generations. Practitioners have long mitigated this with ad hoc deduplication and seed rotation, but no principled framework exists. We introduce Dynamic Context Evolution (DCE), comprising three mechanisms: (1) \emph{verbalized tail sampling} (the model labels each idea with a guess about how obvious it is, and obvious ideas are discarded), which filters high-probability candidates via model self-assessment; (2) \emph{semantic memory}, which maintains a persistent embedding index to reject near-duplicates across batches; and (3) \emph{adaptive prompt evolution}, which reconstructs the generation prompt each batch using memory state and rotating diversity strategies. In experiments across three domains (sustainable packaging concepts, educational exam questions, and creative writing prompts) and two model families (\texttt{gpt-5-mini}\footnote{\texttt{gpt-5-mini-2025-08-07} is a lightweight model in OpenAI's GPT-5 family, accessed via API. It supports structured output. At the time of our experiments, the API did not expose temperature or top-$p$ controls for this snapshot, making diversity interventions at the decoding level unavailable and concept-level approaches like DCE necessary.} and \texttt{claude-haiku-4-5}), A component ablation across 2--3 random seeds per method shows that DCE achieves 0.0 $\pm$ 0.0\% collapse versus 5.6 $\pm$ 2.0\% for naive prompting, while producing 17--18 HDBSCAN clusters per seed versus naive's volatile 2--17, indicating reliably richer conceptual structure. These results are validated with an independent embedding model (\texttt{all-MiniLM-L6-v2}) and hold across sensitivity sweeps of the VTS threshold $\tau$ and dedup threshold $\delta$. Deduplication and prompt evolution are individually insufficient but jointly effective, at approximately \$0.50 per 1{,}000 candidates using only standard API calls, with no fine-tuning or custom architectures required.
\end{abstract}

\section{Introduction}
\label{sec:introduction}

Consider the following experiment. Prompt a language model to generate five educational exam questions. Record the output. Clear the conversation. Repeat 200 times. In the first 30 batches, the model produces genuinely distinct questions spanning diverse topics. By batch 50, the same question structures resurface with superficial variation. By batch 200, 34\% of questions in the final 50 batches are near-duplicates of questions from the first 50, meaning over a third of the output is redundant.

We call this phenomenon \emph{cross-batch mode collapse}: the progressive loss of output diversity when a language model is prompted repeatedly without access to its prior generations. Unlike single-session repetition, which can be mitigated by conversational context, cross-batch collapse is structural. Each API call begins with a blank context window. The model has no mechanism to avoid revisiting high-probability regions of its output distribution. The severity varies by domain (from 4\% in sustainable packaging to 34\% in educational content), but the pattern is consistent: without intervention, repeated prompting converges toward a narrow subset of the output space.

Practitioners have long been aware of this tendency and have adopted ad hoc countermeasures: post-hoc deduplication, seed rotation, prompt paraphrasing, and manual curation. These approaches are effective to varying degrees, but they lack a principled framework: there is no systematic way to know how much diversity is being lost, when additional intervention is needed, or which combination of techniques is sufficient. DCE formalizes these intuitions into a measurable, reproducible pipeline.

This problem is consequential for synthetic data generation, where diversity directly determines downstream utility. A classifier trained on 1{,}000 paraphrases of 50 concepts learns 50 categories. A classifier trained on 1{,}000 genuinely distinct concepts can learn far richer structure.

Existing approaches are insufficient. Token-level interventions such as temperature scaling~\citep{holtzman2020curious} and nucleus sampling introduce lexical variation without altering the underlying concept distribution. Indeed, with \texttt{gpt-5-mini}, these controls are not even available; the API does not expose temperature or top-$p$ parameters. Best-of-$N$ sampling optimizes quality, not diversity. Self-consistency prompting~\citep{wang2023selfconsistency} drives output toward consensus. None provide the model with memory of what it has already generated.

We introduce Dynamic Context Evolution (DCE), a generation framework built on three complementary mechanisms: a self-assessment filter that discards predictable candidates, a persistent semantic memory that prevents near-duplicate acceptance, and an adaptive prompt system that steers each batch toward unexplored conceptual territory. Together, these mechanisms eliminate cross-batch collapse while maintaining output quality. We validate DCE across three domains (sustainable packaging, educational exam questions, and creative writing prompts), two model families (GPT-5-mini and Claude Haiku 4.5), and a seven-method ablation study, demonstrating that the mechanisms are individually insufficient but jointly effective.

\section{Related Work}
\label{sec:related}

\paragraph{Synthetic data generation with LLMs.} Recent work has demonstrated the viability of using language models to generate training data for downstream tasks~\citep{yu2024large, josifoski2023exploiting, li2024synthetic}. However, these approaches typically focus on single-pass generation or quality filtering, and do not address the diversity degradation that occurs across hundreds of independent generation batches.

\paragraph{Model collapse.} \citet{shumailov2024ai} demonstrate that training models on recursively generated synthetic data leads to progressive quality degradation. Our work addresses a related but distinct phenomenon: output repetition within a single generation campaign, where the model's lack of cross-batch memory causes it to revisit high-probability outputs.

\paragraph{Diversity in text generation.} Token-level strategies such as nucleus sampling~\citep{holtzman2020curious} and temperature scaling control the randomness of word selection but do not influence which concepts the model chooses to express. Self-consistency~\citep{wang2023selfconsistency} deliberately reduces diversity by selecting the most common answer across samples. Verbalized probability estimation~\citep{zhang2025verbalized} asks the model to assess the likelihood of its own outputs, which we adapt as a novelty filter.

\paragraph{Active learning and Bayesian optimization.} DCE's exploration-exploitation structure is inspired by active learning~\citep{settles2009active} and Bayesian optimization~\citep{snoek2012practical}, which balance broad search with targeted refinement. DCE adapts this principle informally: early batches explore broadly, later batches target gaps. Unlike these frameworks, DCE does not optimize an explicit acquisition function; the analogy is structural rather than formal.

\paragraph{LLMs as crowdsourcing replacements.} \citet{long2024llms} explore using LLMs to replicate human crowdsourcing pipelines. Our work extends this direction by addressing the specific challenge of maintaining diversity at scale, which human crowds achieve naturally through participant variation but LLMs lack.

\section{Method}
\label{sec:method}

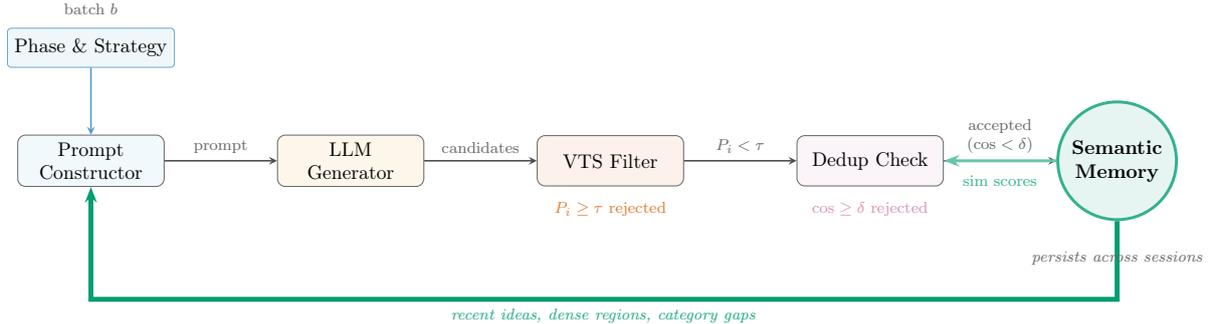
\begin{figure}[h]
    \centering
    \resizebox{\textwidth}{!}{

\begin{tikzpicture}[
    node distance=2.2cm,
    box/.style={rectangle, rounded corners=4pt, draw=black!60, fill=#1!8, minimum width=2.6cm, minimum height=0.9cm, align=center, font=\small},
    box/.default=black,
    flowarrow/.style={->, >=stealth, thick, black!70},
    feedbackarrow/.style={-{Stealth[length=8pt, width=6pt]}, line width=2.5pt, oiteal},
    label/.style={font=\scriptsize, text=black!60},
    phase/.style={rectangle, rounded corners=2pt, draw=oibluishgreen!60, fill=oibluishgreen!6, minimum width=2.2cm, minimum height=0.7cm, align=center, font=\small},
]

\node[box=oiskyblue] (prompt) at (0, 0) {Prompt\\Constructor};
\node[box=oiorange, right=2.0cm of prompt] (gen) {LLM\\Generator};
\node[box=oivermillion, right=2.0cm of gen] (vts) {VTS Filter};
\node[box=oireddishpurple, right=2.0cm of vts] (dedup) {Dedup Check};
\node[circle, draw=oiteal!80, fill=oiteal!10, minimum size=2.0cm, align=center, font=\small\bfseries, line width=1.5pt, right=2.0cm of dedup] (memory) {Semantic\\Memory};

\node[phase, above=1.2cm of prompt] (strategy) {Phase \& Strategy};
\node[label, above=0.1cm of strategy] {batch $b$};
\draw[flowarrow, oibluishgreen!70] (strategy) -- (prompt);

\draw[flowarrow] (prompt) -- node[above, label] {prompt} (gen);
\draw[flowarrow] (gen) -- node[above, label] {candidates} (vts);
\draw[flowarrow] (vts) -- node[above, label] {$P_i < \tau$} (dedup);
\draw[flowarrow] (dedup) -- node[above, label] {\parbox{1.4cm}{\centering\scriptsize accepted\\($\cos < \delta$)}} (memory);

\node[label, below=0.15cm of vts] {\textcolor{oivermillion!80}{$P_i \geq \tau$ rejected}};
\node[label, below=0.15cm of dedup] {\textcolor{oireddishpurple!80}{$\cos \geq \delta$ rejected}};

\draw[feedbackarrow, oiteal!60, line width=1.5pt] (memory) -- node[below, label, yshift=-0.1cm] {\textcolor{oiteal!80}{sim scores}} (dedup);

\draw[feedbackarrow, oiteal]
    (memory.south) -- ++(0, -1.4)
    -| node[pos=0.25, below, label] {\textcolor{oiteal!90}{\textit{recent ideas, dense regions, category gaps}}}
    (prompt.south);

\node[label, font=\scriptsize\itshape, below=0.4cm of memory] {persists across sessions};

\end{tikzpicture}}
    \caption{System architecture of the DCE generation loop. The pipeline flows left to right: the prompt constructor assembles each batch's prompt from phase/strategy settings and semantic memory state, the LLM generates candidates, and successive filters (VTS, dedup) gate acceptance into the memory bank. The feedback loop along the bottom returns memory state (recent ideas, dense regions, category gaps) to the prompt constructor for the next batch.}
    \label{fig:architecture}
\end{figure}

DCE augments the standard generation pipeline with three mechanisms, each targeting a specific failure mode of repeated independent prompting. Figure~\ref{fig:architecture} shows the complete generation loop, and Figure~\ref{fig:edv_concept} illustrates how the filtering mechanisms interact to select candidates.

\subsection{Verbalized Tail Sampling}
\label{sec:vts}

Each candidate idea is generated alongside a self-assessed probability score: the model labels each idea with a guess about how obvious it is, and obvious ideas are discarded. Specifically, the model estimates the likelihood that another language model, given the same domain prompt, would produce the same concept. Candidates with probability estimates above a threshold $\tau = 0.10$ are discarded.

For example, a candidate ``smart water bottle'' with $P = 0.45$ is rejected as predictable, while ``shipping containers with walls of compressed agricultural waste that decompose into fertilizer'' with $P = 0.03$ is retained.

The probability estimates are not calibrated in an absolute sense~\citep{zhang2025verbalized}. However, they provide a reliable relative ordering between obvious and non-obvious candidates, which is sufficient for filtering purposes. This mechanism targets the \emph{depth} dimension of diversity: it ensures that accepted ideas are not merely the most probable completions.

\subsection{Semantic Memory}
\label{sec:memory}

Every accepted idea is embedded into a 1536-dimensional vector using \texttt{text\allowbreak-embedding\allowbreak-3\allowbreak-small}~\citep{neelakantan2022text} and stored in a persistent vector database (ChromaDB~\citep{chromadb2023}). Before a new candidate is accepted, the system computes its cosine similarity to all stored embeddings. Any candidate with similarity exceeding $\delta = 0.85$ to any stored idea is rejected as a near-duplicate.

This mechanism operates at the semantic level rather than the lexical level. ``Smart water bottle'' and ``intelligent hydration vessel'' share few words but produce nearly identical embeddings; the memory filter catches both. The database persists across sessions, ensuring that generation can be paused and resumed without loss of deduplication state.

This mechanism targets the \emph{breadth} dimension of diversity: it prevents the accepted set from accumulating redundant entries, even when the redundancy is expressed through different surface forms.

\subsection{Adaptive Prompt Evolution}
\label{sec:prompt_evolution}

The generation prompt is reconstructed before each batch using three sources of information derived from the current memory state:

\begin{itemize}
    \item The 10 most recently accepted ideas, presented as exclusions
    \item 5 ideas from the densest regions of the embedding space, flagged as over-represented territory
    \item The current distribution of ideas across categories, highlighting underrepresented areas
\end{itemize}

In addition, each batch applies one of four diversity strategies in round-robin rotation:

\begin{description}
    \item[Gap targeting.] Directs the model toward categories with the fewest accepted ideas, explicitly prioritizing underrepresented areas of the concept space.
    \item[Assumption inversion.] Identifies assumptions implicit in recent ideas and instructs the model to invert them, forcing exploration of complementary regions.
    \item[Cross-industry stimulus.] Provides analogies from unrelated domains (e.g., hospitality, aerospace, marine engineering) to break domain-specific fixation. Combining ideas from distant fields produces concepts that within-domain prompting rarely generates.
    \item[Constraint variation.] Applies extreme constraints (e.g., ``must cost nothing,'' ``must work without electricity,'' ``must be reusable 100 times'') to shift the model away from default design assumptions.
\end{description}

\paragraph{Category quality.} Gap targeting relies on the model's self-assigned category labels to identify underrepresented areas. The consistency of these labels is validated empirically in Section~\ref{sec:category_quality}.

The generation campaign is divided into two phases: the first 40\% of batches prioritize broad exploration, while the remaining 60\% shift to gap-filling within the discovered space.

\paragraph{Prompt assembly example.} Figure~\ref{fig:prompt_example} shows a simplified version of the prompt sent at batch 100 (gap-targeting strategy, exploitation phase). The prompt integrates memory state (recent ideas, category distribution, dense regions) with the current strategy directive, giving the model specific guidance on where to explore next.

\begin{figure}[h]
\centering
\begin{lstlisting}[
  basicstyle=\ttfamily\scriptsize,
  frame=single,
  xleftmargin=0.5em,
  breaklines=true,
  escapeinside={(*}{*)},
  columns=fullflexible,
]
Generate 5 sustainable packaging concepts.     (*\textrm{\scriptsize\color{black!50} $\leftarrow$ base template}*)

For each idea, estimate the probability that
another LLM would generate the same concept.   (*\textrm{\scriptsize\color{black!50} $\leftarrow$ VTS instruction}*)

PHASE: Exploitation (gap-filling).             (*\textrm{\scriptsize\color{black!50} $\leftarrow$ from batch number}*)
Focus on filling gaps in underrepresented
categories rather than broad exploration.

STRATEGY: Gap targeting.                       (*\textrm{\scriptsize\color{black!50} $\leftarrow$ round-robin}*)
The following categories have the fewest
accepted ideas. Prioritize these areas:
  - Thermal regulation (2 ideas)
  - Ocean-degradable materials (3 ideas)
  - Agricultural waste reuse (4 ideas)

RECENT IDEAS (do not repeat these):            (*\textrm{\scriptsize\color{black!50} $\leftarrow$ from memory}*)
  - Kelp-fiber insulation panels
  - Phase-change cooling wraps
  - ... (10 most recent)

OVER-REPRESENTED (avoid these regions):        (*\textrm{\scriptsize\color{black!50} $\leftarrow$ from memory}*)
  - Biodegradable films (dense cluster)
  - Mushroom-based materials (dense cluster)

CATEGORY DISTRIBUTION:                         (*\textrm{\scriptsize\color{black!50} $\leftarrow$ from memory}*)
  Biodegradable films: 47 | Reusable: 38
  Edible packaging: 31   | Thermal: 2  ...
\end{lstlisting}
\caption{Simplified prompt at batch 100 (exploitation phase, gap-targeting strategy). Annotations show which DCE component contributes each section. The full prompt template is in Appendix~\ref{sec:appendix_prompts}.}
\label{fig:prompt_example}
\end{figure}

\paragraph{Prompt evolution over time.} Figure~\ref{fig:prompt_evolution} contrasts the prompt at batch 10 (exploration phase, sparse memory) with batch 190 (exploitation phase, full memory), showing how the prompt adapts as the concept space fills.

\begin{figure}[h]
\centering
\begin{minipage}[t]{0.48\textwidth}
\begin{lstlisting}[
  basicstyle=\ttfamily\tiny,
  frame=single,
  xleftmargin=0.3em,
  breaklines=true,
  escapeinside={(*}{*)},
  columns=fullflexible,
  title={\scriptsize\textbf{Batch 10} (exploration, cross-industry)}
]
PHASE: Exploration.
Focus on breadth -- explore widely.

STRATEGY: Cross-industry stimulus.
Imagine packaging designed by someone
from: Veterinary medicine, Textile
manufacturing, Urban planning.

RECENT IDEAS (do not repeat):
  - Mycelium foam inserts
  - Edible rice-paper wraps
  ... (8 more, all from batches 1-9)

CATEGORY DISTRIBUTION (3 total):
  Biodegradable films: 12
  Edible packaging: 8
  Reusable containers: 7
  (*\textit{(only 3 categories after 10 batches)}*)
\end{lstlisting}
\end{minipage}
\hfill
\begin{minipage}[t]{0.48\textwidth}
\begin{lstlisting}[
  basicstyle=\ttfamily\tiny,
  frame=single,
  xleftmargin=0.3em,
  breaklines=true,
  escapeinside={(*}{*)},
  columns=fullflexible,
  title={\scriptsize\textbf{Batch 190} (exploitation, gap targeting)}
]
PHASE: Exploitation (gap-filling).
Focus on filling underrepresented areas.

STRATEGY: Gap targeting.
These categories need more ideas:
  - Thermal regulation (2 ideas)
  - Ocean-degradable materials (3 ideas)
  - Agricultural waste reuse (4 ideas)

RECENT IDEAS (do not repeat):
  - Compressed hemp fiber crates
  - Algae-ink printed labels
  ... (8 more, from batches 180-189)

OVER-REPRESENTED (avoid):
  - Biodegradable films (dense cluster)
  - Mushroom-based materials (dense)
  - Edible coatings (dense cluster)

CATEGORY DISTRIBUTION (47 total):
  Biodegradable films: 47
  Reusable containers: 38
  Edible packaging: 31
  Mycelium-based: 28
  Seaweed/algae: 22
  Agricultural waste: 16
  Insulation/thermal: 12
  Water-soluble: 9
  Compressed fiber: 7
  Mineral-based: 5
  Thermal regulation: 2
  Ocean-degradable: 3
  (*\textit{(+ 35 more categories with 1--4 ideas)}*)
\end{lstlisting}
\end{minipage}

\vspace{0.5em}
\centering
\begin{minipage}[t]{0.48\textwidth}
\centering\small
\textcolor{black!60}{\rule{3cm}{0.4pt}}\\[2pt]
\textbf{3} categories $\cdot$ \textbf{27} ideas $\cdot$ exploration phase
\end{minipage}
\hfill
\begin{minipage}[t]{0.48\textwidth}
\centering\small
\textcolor{black!60}{\rule{3cm}{0.4pt}}\\[2pt]
\textbf{47} categories $\cdot$ \textbf{950} ideas $\cdot$ 3 dense regions flagged
\end{minipage}

\caption{Prompt evolution: batch 10 (left) vs.\ batch 190 (right). The summary callouts below each panel highlight the key quantitative difference: early prompts have sparse memory and prioritize broad exploration, while late prompts have rich category distributions, identify dense regions to avoid, and target specific gaps.}
\label{fig:prompt_evolution}
\end{figure}

\paragraph{Strategy examples.} Figure~\ref{fig:strategy_examples} illustrates all four diversity strategies with concrete prompt snippets and example outputs.

\begin{figure}[h]
\centering
\begin{lstlisting}[
  basicstyle=\ttfamily\scriptsize,
  frame=single,
  xleftmargin=0.5em,
  breaklines=true,
  escapeinside={(*}{*)},
  columns=fullflexible,
]
STRATEGY: Gap targeting.                       (*\textrm{\scriptsize\color{black!50} $\leftarrow$ round-robin}*)
These categories are underrepresented:
  - Thermal regulation (2 ideas)
  - Ocean-degradable materials (3 ideas)
At least half your ideas MUST target these.
(*\textrm{\scriptsize\color{oiteal} \textbf{Output:} Seaweed-cellulose thermal liner}*)
(*\textrm{\scriptsize\color{oiteal} that regulates temperature via phase change}*)
(*\textrm{\scriptsize\color{oiteal} and dissolves in seawater (P = 0.03).}*)
-------------------------------------------
STRATEGY: Assumption inversion.                (*\textrm{\scriptsize\color{black!50} $\leftarrow$ round-robin}*)
Recent ideas assume the following:
  - Packaging is disposable after use
  - Consumer removes packaging manually
Invert each assumption. Generate ideas where:
  - Packaging is permanent / reusable
  - Packaging removes itself (dissolves, etc.)
(*\textrm{\scriptsize\color{oiteal} \textbf{Output:} Self-dissolving mineral shell}*)
(*\textrm{\scriptsize\color{oiteal} that liquefies on contact with water,}*)
(*\textrm{\scriptsize\color{oiteal} leaving zero waste (P = 0.04).}*)
-------------------------------------------
STRATEGY: Cross-industry stimulus.             (*\textrm{\scriptsize\color{black!50} $\leftarrow$ round-robin}*)
Imagine packaging designed by someone from:
  - Aerospace engineering
  - Marine biology
What would they do differently?
(*\textrm{\scriptsize\color{oiteal} \textbf{Output:} Pressure-sealed ablative wrap}*)
(*\textrm{\scriptsize\color{oiteal} (aerospace) that sheds layers in transit,}*)
(*\textrm{\scriptsize\color{oiteal} each composting independently (P = 0.02).}*)
-------------------------------------------
STRATEGY: Constraint variation.                (*\textrm{\scriptsize\color{black!50} $\leftarrow$ round-robin}*)
CONSTRAINT: The packaging must cost nothing
to produce (use only waste materials).
(*\textrm{\scriptsize\color{oiteal} \textbf{Output:} Compressed coffee-ground bricks}*)
(*\textrm{\scriptsize\color{oiteal} collected from cafes, molded into shipping}*)
(*\textrm{\scriptsize\color{oiteal} containers that compost in 30 days (P = 0.02).}*)
\end{lstlisting}
\caption{All four diversity strategies with prompt snippets and example outputs. Gap targeting steers toward underrepresented categories. Assumption inversion negates implicit assumptions in recent ideas. Cross-industry stimulus introduces framing from distant domains. Constraint variation applies extreme constraints to break default design assumptions.}
\label{fig:strategy_examples}
\end{figure}

\subsection{Generation Loop}

Algorithm~\ref{alg:dce} summarizes the complete DCE pipeline.

\begin{algorithm}[h]
\caption{Dynamic Context Evolution}
\label{alg:dce}
\begin{algorithmic}[1]
\For{each batch $b = 1$ to $B$}
    \State Construct prompt from memory state, strategy $s = b \bmod 4$, and phase
    \State Generate $n$ candidate ideas with structured output
    \State Filter: discard candidates with self-assessed $P_i \geq \tau$
    \State Deduplicate: discard candidates with $\max_{j \in M} \cos(\mathbf{e}_i, \mathbf{e}_j) \geq \delta$
    \State Update memory: embed and store all accepted candidates
\EndFor
\end{algorithmic}
\end{algorithm}

\noindent where $\tau = 0.10$ is the probability threshold and $\delta = 0.85$ is the similarity threshold.

\section{Evaluation Metrics}
\label{sec:metrics}

Standard text generation metrics (perplexity, BLEU, ROUGE) measure quality or fidelity, not conceptual diversity across a large set of generated items. We introduce Effective Diversity Volume (EDV) and use it alongside collapse rate to evaluate sustained diversity.

\subsection{Effective Diversity Volume}

EDV measures whether each batch continues to produce ideas that are both \emph{surprising} (low model-assessed probability) and \emph{novel} (far from all previously accepted ideas). For a batch of $n$ accepted ideas:

\begin{equation}
\text{EDV}_{\text{batch}} = \frac{1}{n} \sum_{i=1}^{n} \underbrace{(1 - P_i)}_{\text{depth}} \times \underbrace{\min_{j \in M} (1 - \cos(\mathbf{e}_i, \mathbf{e}_j))}_{\text{breadth}}
\end{equation}

\noindent where $P_i$ is the self-assessed probability and $M$ is the memory bank at the time of evaluation. A flat EDV curve over batches indicates sustained exploration; a declining curve indicates convergence toward previously visited regions.

\paragraph{Why multiplication?} The multiplicative form is deliberate: it requires both depth \emph{and} breadth for a high score. An idea that is surprising but nearly identical to a stored idea ($P_i = 0.02$, breadth $\approx 0$) contributes near-zero EDV, as does an obvious but distant idea ($P_i = 0.90$, breadth $= 0.5$). A weighted sum would allow one strong dimension to compensate for a weak one, which would not reflect our design goal: we want ideas that are \emph{both} non-obvious and semantically novel. The multiplicative form encodes this conjunctive requirement directly. We note that this encodes a specific design preference, not a universal truth about diversity. In domains where novel twists on familiar concepts matter more than entirely new territory, an additive form may be preferable.

To verify that our results are not an artifact of the multiplicative form, Table~\ref{tab:edv_formulations} compares method rankings under all three formulations using the packaging domain data (seed 42, \texttt{all-MiniLM-L6-v2} embeddings).

\begin{table}[h]
\centering
\caption{EDV retention (\%) under three formulations (packaging domain, seed 42, MiniLM embeddings). Rankings are nearly identical: VTS + dedup is the lowest-ranked method in all three, and the top methods cluster within a narrow range. Minor rank swaps occur at the top (naive and prompt evo + dedup exchange positions 1--2 under the additive form), but the overall ordering is stable.}
\label{tab:edv_formulations}
\begin{tabular}{lcccc}
\toprule
Method & Multiplicative & Additive & Geometric & Rank \\
\midrule
Naive & 32.7 & 64.0 & 56.9 & 1 / 2 / 1 \\
Prompt evo + dedup & 31.4 & 64.9 & 55.9 & 2 / 1 / 2 \\
DCE (full) & 28.7 & 63.7 & 53.4 & 3 / 3 / 3 \\
VTS only & 28.0 & 62.7 & 52.8 & 4 / 4 / 4 \\
Dedup only & 27.2 & 62.3 & 52.0 & 5 / 5 / 5 \\
Prompt evo only & 24.5 & 61.4 & 49.2 & 6 / 6 / 6 \\
VTS + dedup & 19.7 & 58.5 & 43.9 & 7 / 7 / 7 \\
\bottomrule
\end{tabular}
\end{table}

\noindent Rankings are stable across formulations (ranks 3--7 are identical; only the top two swap under additive). Under all three formulations, naive ranks first or second in EDV retention. This reflects a property of the retention ratio: naive starts with a lower absolute EDV (its early batches are unconstrained and initially diverse), so the ratio of late-to-early EDV can be higher even when absolute late-batch diversity is worse. For absolute EDV values and collapse rates, which together give the full picture, see Table~\ref{tab:exp2}. We retain the multiplicative form for its interpretive clarity as a conjunctive requirement.

We report both absolute EDV values and the retention ratio (EDV at batch 200 divided by EDV at batch 1) throughout. The retention ratio captures how well diversity is maintained as the concept space fills, while absolute values reveal the starting level. A method could maintain high retention of a low initial EDV, which the absolute values expose.

\paragraph{Informal calibration.} An EDV retention of 23.6\% means that batch 200 ideas score roughly one-quarter as high on the depth$\times$breadth product as batch 1 ideas, but still pass both filters (0\% collapse rate at $\delta = 0.85$). Late-batch ideas occupy denser but non-overlapping regions of the concept space: they are less surprising individually but remain semantically distinct from all prior ideas.

\begin{figure}[h]
    \centering

\begin{tikzpicture}[
    cell/.style={rectangle, rounded corners=4pt, minimum width=5.6cm, minimum height=3cm, align=center, font=\small, text width=5cm, inner sep=8pt},
    elabel/.style={font=\footnotesize, text=black!60},
]

\fill[oiteal!8] (0, 0) rectangle (6, 3.4);         
\fill[oivermillion!6] (-6, 0) rectangle (0, 3.4);   
\fill[oiorange!6] (0, -3.4) rectangle (6, 0);       
\fill[oireddishpurple!6] (-6, -3.4) rectangle (0, 0); 

\draw[black!25, thick] (-6, 0) -- (6, 0);
\draw[black!25, thick] (0, -3.4) -- (0, 3.4);

\node[font=\small\bfseries, anchor=south] at (0, 3.7) {Breadth (distance from memory)};
\node[font=\small\bfseries, anchor=south, rotate=90] at (-6.8, 0) {Depth (how surprising)};

\node[elabel, anchor=east] at (-0.2, 2.8) {high};
\node[elabel, anchor=east] at (-0.2, -2.8) {low};
\node[elabel, anchor=north] at (-3.2, -0.15) {low};
\node[elabel, anchor=north] at (3.2, -0.15) {high};

\node[cell, fill=oiteal!10, draw=oiteal!50] at (3, 1.7) {
    \textbf{\textcolor{oiteal}{Accepted (high EDV)}}\\[3pt]
    surprising \emph{and} far from memory\\[5pt]
    \textcolor{black!60}{\emph{``walls made of compressed}}\\[-1pt]
    \textcolor{black!60}{\emph{agricultural waste''}}\\[2pt]
    \textcolor{black!60}{$P = 0.03$\enspace$\cdot$\enspace nearest = 0.31}
};

\node[cell, fill=oivermillion!8, draw=oivermillion!40] at (-3, 1.7) {
    \textbf{\textcolor{oivermillion}{Rejected by dedup}}\\[3pt]
    surprising but too similar\\
    to something in memory\\[5pt]
    \textcolor{black!60}{\emph{``intelligent hydration vessel''}}\\[2pt]
    \textcolor{black!60}{$P = 0.05$\enspace$\cdot$\enspace nearest = 0.92}
};

\node[cell, fill=oiorange!8, draw=oiorange!40] at (3, -1.7) {
    \textbf{\textcolor{oiorange}{Rejected by VTS}}\\[3pt]
    novel territory but obvious idea\\[5pt]
    \textcolor{black!60}{\emph{``smart water bottle''}}\\[2pt]
    \textcolor{black!60}{$P = 0.45$\enspace$\cdot$\enspace nearest = 0.22}
};

\node[cell, fill=oireddishpurple!10, draw=oireddishpurple!40] at (-3, -1.7) {
    \textbf{\textcolor{oireddishpurple}{Rejected by both}}\\[3pt]
    obvious \emph{and} redundant\\[5pt]
    \textcolor{black!60}{\emph{``eco-friendly water bottle''}}\\[2pt]
    \textcolor{black!60}{$P = 0.52$\enspace$\cdot$\enspace nearest = 0.91}
};

\node[font=\small, anchor=north] at (0, -4.0) {
    $\text{EDV}_{\text{batch}} = \frac{1}{n}\sum_i \text{depth}_i \times \text{breadth}_i$
    \quad\textcolor{black!50}{(averaged over accepted ideas per batch)}
};

\end{tikzpicture}
    \caption{The EDV filtering space. Each candidate is scored on depth (how surprising, determined by VTS) and breadth (how distant from memory, determined by deduplication). Only candidates that pass both filters (top-right quadrant) contribute to EDV. The quadrant labels show which filter rejects candidates in each region.}
    \label{fig:edv_concept}
\end{figure}

\subsection{Collapse Rate}

Collapse rate measures the proportion of ideas in the final 50 batches that are near-duplicates (cosine similarity $> 0.85$) of ideas from the first 50 batches. A collapse rate of 0\% indicates that the end of the generation campaign is as conceptually original as the beginning. We distinguish the underlying phenomenon (the model revisiting high-probability outputs across independent sessions) from this operationalization, which uses cosine similarity exceeding $\delta$ as a proxy. The measured collapse rate depends on both the embedding model and the threshold; our cross-embedding validation (Section~\ref{sec:edv_validation}) confirms that the results are robust to these choices.

\section{Experiments}
\label{sec:experiments}

Unless otherwise noted, experiments use \texttt{gpt\allowbreak-5\allowbreak-mini\allowbreak-2025\allowbreak-08\allowbreak-07} for generation and \texttt{text\allowbreak-embedding\allowbreak-3\allowbreak-small} (1536 dimensions) for embeddings, generating ideas in batches of~5 across 200 batches (1{,}000 candidate ideas per method; accepted idea counts vary by method due to filtering). We evaluate on three domains: sustainable packaging concepts, educational exam questions for introductory biology and computer science, and creative writing prompts.

\subsection{Experiment 1: Characterizing Cross-Batch Collapse}
\label{sec:exp1}

We first establish that cross-batch collapse is a real and measurable phenomenon under naive prompting (independent batches with no memory or filtering).

\begin{figure}[h]
    \centering
    \input{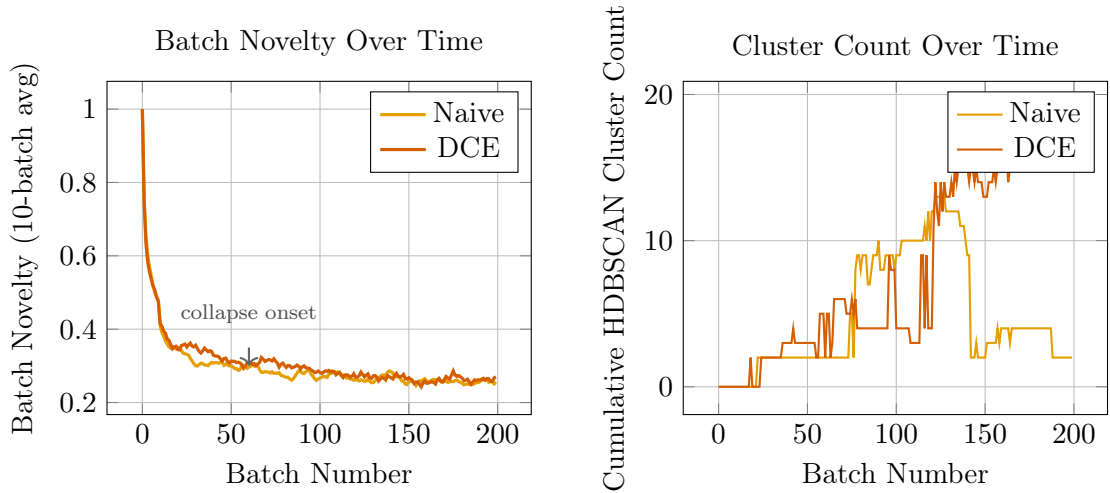}
    \caption{Collapse characterization: naive prompting vs.\ DCE (200 batches, 10-batch rolling averages). Left: naive novelty decays steadily after batch ${\sim}$60 (annotated), while DCE maintains elevated novelty throughout. Right: cluster count saturates early under naive prompting but continues growing under DCE, indicating sustained conceptual exploration.}
    \label{fig:collapse_comparison}
\end{figure}

Figure~\ref{fig:collapse_comparison} presents two complementary views, comparing naive prompting against DCE. Under naive prompting, batch novelty (average minimum cosine distance between new ideas and all prior ideas) drops from 1.0 to approximately 0.20 within 60 batches. Concurrently, HDBSCAN~\citep{campello2013hdbscan} cluster count plateaus after batch 60, indicating that subsequent batches contribute no new conceptual categories. DCE maintains substantially higher novelty throughout and continues growing the cluster count past batch 100.

\subsection{Experiment 2: Component Ablation}
\label{sec:exp2}

We compare seven configurations to isolate the contribution of each mechanism. The first four represent the original ablation; the remaining three decompose DCE's components further:

\begin{enumerate}
    \item Naive: independent batches, no filtering or memory
    \item VTS only: verbalized tail sampling filter, no memory or prompt evolution
    \item VTS + dedup: VTS with semantic memory, no prompt evolution
    \item DCE (full): all three mechanisms (VTS + memory + adaptive prompt evolution)
    \item Dedup only: semantic memory without VTS or prompt evolution
    \item Prompt evo only: adaptive prompts without VTS or dedup
    \item Prompt evo + dedup: adaptive prompts with semantic memory, no VTS
\end{enumerate}

Note that token-level baselines (temperature scaling, nucleus sampling) were planned but could not be executed because \texttt{gpt-5-mini} does not expose temperature or top-$p$ parameters; only the default decoding configuration is available.

\begin{table}[h]
\centering
\caption{Ablation results on the packaging domain across 200 batches. The top four methods report mean $\pm$ std over 2--3 random seeds (see Seeds column); ablation-only methods (below the rule) use a single seed. EDV retention is the ratio of last-batch to first-batch EDV. Methods using dedup achieve 0\% collapse consistently; methods without it do not. DCE achieves both the highest mean EDV retention and zero collapse across all seeds.}
\label{tab:exp2}
\begin{tabular}{lccc}
\toprule
Method & Seeds & EDV Retention & Collapse \\
\midrule
\multicolumn{4}{l}{\textit{Multi-seed methods (mean $\pm$ std)}} \\[2pt]
Naive & 3 & 20.4 $\pm$ 6.1\% & 5.6 $\pm$ 2.0\% \\
VTS only & 2 & 22.4 $\pm$ 1.3\% & 3.2 $\pm$ 1.2\% \\
VTS + dedup & 3 & 22.1 $\pm$ 2.7\% & 0.0 $\pm$ 0.0\% \\
DCE (full) & 3 & 23.6 $\pm$ 3.5\% & 0.0 $\pm$ 0.0\% \\
\midrule
\multicolumn{4}{l}{\textit{Single-seed ablations}} \\[2pt]
Dedup only & 1 & 26.1\% & 0.0\% \\
Prompt evo only & 1 & 24.7\% & 2.8\% \\
Prompt evo + dedup & 1 & 25.8\% & 0.0\% \\
Seed rotation + dedup (3 seeds) & 1 & 21.6\% & 0.0\% \\
\bottomrule
\end{tabular}
\end{table}

The results in Table~\ref{tab:exp2}, reported as mean $\pm$ std over 2--3 seeds per method, reveal several patterns. The primary result is collapse rate: any method that includes deduplication achieves 0.0 $\pm$ 0.0\% collapse across all seeds, while methods without it show nonzero collapse (naive: 5.6 $\pm$ 2.0\%), confirming that semantic memory is necessary for reliable collapse prevention. Second, HDBSCAN cluster analysis (Table~\ref{tab:cluster_counts}) shows that DCE produces 17--18 clusters consistently across seeds, versus naive's highly variable 2--17, demonstrating that DCE reliably produces richer conceptual structure. Third, DCE achieves the highest mean EDV retention (23.6 $\pm$ 3.5\%), though with only 3 seeds this difference is not statistically significant ($p = 0.50$, Wilcoxon signed-rank) and should be interpreted as directional evidence, not a confirmed effect. The full DCE pipeline's advantage comes from the prompt actively steering the model toward underexplored territory each batch; adding deduplication to VTS without prompt evolution yields comparable EDV retention (22.1 $\pm$ 2.7\%) because the memory filter rejects near-duplicates without providing alternative directions. Ablation-only methods (below the rule in Table~\ref{tab:exp2}) report single-seed results; in particular, dedup-only's 26.1\% EDV retention exceeds DCE's multi-seed mean of 23.6\%, and this comparison should be interpreted cautiously given the high variance across seeds.

\paragraph{Seed rotation baseline.} A common practitioner approach is to rotate across multiple random seeds and deduplicate the pooled output. To test this, we pool naive generations from three seeds (42, 123, 456) and process them in round-robin order: all ideas from seed 42 batch 1, then seed 123 batch 1, then seed 456 batch 1, then seed 42 batch 2, and so on. We then apply greedy post-hoc deduplication at $\delta = 0.85$, processing ideas in this interleaved order. Greedy dedup is order-dependent (the first idea encountered is kept; later near-duplicates are rejected), so the interleaving simulates the round-robin seed rotation that practitioners commonly use. The result (bottom row of Table~\ref{tab:exp2}) shows that seed rotation with dedup achieves 0\% collapse and 21.6\% EDV retention but, despite starting from 3$\times$ the candidate pool (3{,}010 ideas from 3 seeds), produces lower EDV retention than DCE's multi-seed mean of 23.6\% (Table~\ref{tab:exp2}). Seed rotation increases the initial pool through model stochasticity, but without adaptive steering, the model exhausts its repertoire in each seed independently, and post-hoc dedup can only select from what was generated. DCE's prompt evolution actively steers toward underexplored territory each batch, achieving higher EDV retention from fewer total candidates.

\paragraph{Why not just deduplicate?} Dedup-only achieves 0\% collapse and 26.1\% EDV retention (single seed), which exceeds DCE's multi-seed mean of 23.6\%. However, prompt evolution's key advantage is conceptual breadth, not just collapse prevention: DCE produces 18 HDBSCAN clusters by batch 200 versus 9 for dedup-only (Table~\ref{tab:cluster_counts}). Dedup-only narrows the accepted set to whatever the model happens to generate without redirection; prompt evolution actively steers toward underexplored territory.

\begin{figure}[h]
    \centering
    \includegraphics[width=\textwidth]{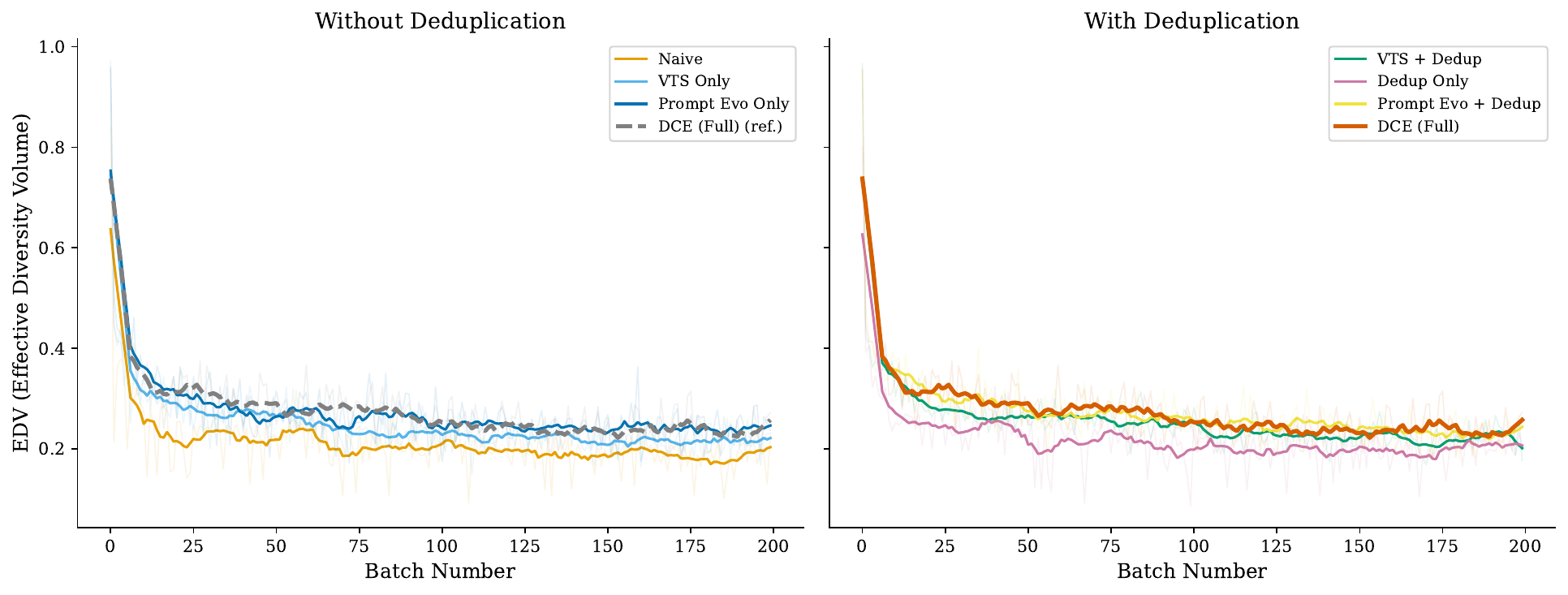}
    \caption{EDV over 200 batches (10-batch rolling averages; raw values shown faintly), split by deduplication status. Left: methods without dedup all show declining EDV; the smoothed DCE reference (gray dashed) is clearly above all of them by batch 50. Right: methods with dedup achieve 0\% collapse; DCE (thick line) maintains the highest EDV throughout, confirming that prompt evolution adds value beyond dedup alone.}
    \label{fig:edv}
\end{figure}

Figure~\ref{fig:edv} shows the EDV trajectory for all seven methods across 200 batches, split into two panels for readability: methods without deduplication (left) and methods with deduplication (right). The DCE curve appears as a gray dashed reference in the left panel.

Table~\ref{tab:cluster_counts} reports HDBSCAN cluster counts across all methods, computed using an independent embedding model (\texttt{all-MiniLM-L6-v2}) for consistency. By batch 200, DCE and prompt-evolution methods produce the richest cluster structure (17--19 clusters), far exceeding naive's 2. At earlier checkpoints individual methods occasionally exceed DCE (e.g., dedup only at batch 50, prompt evo + dedup at batch 100), but these intermediate counts reflect transient embedding geometry rather than stable conceptual breadth (see below).

\begin{table}[h]
\centering
\caption{HDBSCAN cluster counts at batch milestones (MiniLM embeddings, \texttt{min\_cluster\_size}${}=5$). By batch 200, DCE and methods with prompt evolution maintain 17--19 clusters, while naive generation collapses to just 2. Cluster counts are sensitive to \texttt{min\_cluster\_size}: at $\{5, 7, 10\}$ DCE produces more clusters than naive, but at $\texttt{min\_cluster\_size}{=}3$ naive exceeds DCE (32.7 vs.\ 19.7), likely because fine-grained clustering fragments naive's repetitive output into many small groups. Absolute counts across $\{3, 5, 7, 10\}$: DCE 19.7, 17.7, 11.7, 5.0; naive 32.7, 11.0, 8.0, 4.3. Across seeds 42, 123, 456, DCE produces 18, 18, 17 clusters respectively (naive: 2, 14, 17), with DCE showing notably lower variance.}
\label{tab:cluster_counts}
\begin{tabular}{lccc}
\toprule
Method & Clusters@50 & Clusters@100 & Clusters@200 \\
\midrule
Naive & 2 & 8 & 2 \\
VTS only & 2 & 6 & 14 \\
VTS + dedup & 5 & 5 & 2 \\
DCE (full) & 3 & 8 & 18 \\
\midrule
Dedup only & 5 & 2 & 9 \\
Prompt evo only & 3 & 8 & 17 \\
Prompt evo + dedup & 2 & 12 & 19 \\
\midrule
\multicolumn{4}{l}{\textit{Cross-seed summary at batch 200 (seeds 42, 123, 456)}} \\[2pt]
DCE (3 seeds) & --- & --- & 17.7 $\pm$ 0.6 \\
Naive (3 seeds) & --- & --- & 11.0 $\pm$ 7.9 \\
\bottomrule
\end{tabular}
\end{table}

\paragraph{Interpreting intermediate cluster counts.} The non-monotonic trajectories in Table~\ref{tab:cluster_counts} (e.g., VTS + dedup at 5$\to$5$\to$2, dedup-only at 5$\to$2$\to$9) reflect transient embedding geometry, not stable measures of conceptual breadth. HDBSCAN is density-sensitive: adding ideas can merge or split clusters as local density changes. The batch-200 counts, aggregated over seeds, are the appropriate comparison point. DCE's cross-seed consistency at batch 200 (18, 18, 17 clusters) versus naive's volatility (2, 14, 17) is itself informative: DCE reliably produces a rich cluster structure regardless of random seed. Intermediate checkpoints are included for completeness but should not be over-interpreted.

\subsection{EDV Validation with Independent Embeddings}
\label{sec:edv_validation}

Since both EDV components rely on pipeline-internal signals, we validate the metric using an independent embedding model. We take the same generated outputs from the packaging domain and re-embed them with \texttt{all-MiniLM-L6-v2}~\citep{reimers2019sentencebert} (384 dimensions, trained on a different corpus than \texttt{text-embedding-3-small}), then recompute EDV retention and collapse rate. No new generation is performed; only the embedding step differs.

\begin{table}[h]
\centering
\caption{EDV validation: original OpenAI embeddings vs.\ independent MiniLM embeddings (seed 42; Table~\ref{tab:exp2} reports multi-seed means). The bottom two ranks are stable across embeddings (VTS + dedup lowest, VTS only third), but DCE and naive swap at the top: DCE leads under OpenAI embeddings while naive leads under MiniLM.}
\label{tab:edv_validation}
\begin{tabular}{lcc}
\toprule
Method & EDV ret (OpenAI) & EDV ret (MiniLM) \\
\midrule
Naive & 27.0\% & 32.7\% \\
VTS only & 23.8\% & 28.0\% \\
VTS + dedup & 18.3\% & 19.7\% \\
DCE (full) & 28.5\% & 28.7\% \\
\bottomrule
\end{tabular}
\end{table}

Table~\ref{tab:edv_validation} shows partial rank stability across embeddings. The bottom two ranks are preserved (VTS + dedup lowest, VTS only third), but the top two methods swap: DCE leads under OpenAI embeddings (28.5\% vs.\ 27.0\% for naive), while naive leads under MiniLM (32.7\% vs.\ 28.7\% for DCE). This swap is consistent with naive's higher absolute retention under MiniLM across all methods, suggesting that the MiniLM embedding space compresses the distances that separate naive ideas less aggressively. The key structural finding --- that dedup-enabled methods occupy distinct rank positions from non-dedup methods --- holds under both embeddings.

Collapse rate robustness was also tested with MiniLM embeddings at three thresholds: at $\delta = 0.80$, 0.85, and 0.90, DCE and VTS + dedup maintain 0\% collapse, while naive shows 0.8\% at $\delta = 0.85$. This confirms that the collapse-prevention results are not an artifact of the generation embedding model.

\subsection{VTS Rejection Analysis}
\label{sec:vts_analysis}

To understand what VTS actually filters, we analyze the naive generation data (which contains no filtering) as a proxy for the full generation distribution. For each naive idea, we determine whether VTS would reject it ($P_i \geq 0.10$) and whether dedup would reject it (cosine similarity $> 0.85$ to any prior idea), producing a 2$\times$2 confusion matrix.

\begin{table}[h]
\centering
\caption{VTS vs.\ dedup decision analysis on naive generation data (1{,}000 ideas). VTS and dedup target largely non-overlapping populations: 96.9\% of VTS-rejected ideas are semantically novel (would pass dedup), confirming that VTS filters on predictability, not redundancy.}
\label{tab:vts_confusion}
\begin{tabular}{lcc}
\toprule
 & Dedup accept & Dedup reject \\
\midrule
VTS accept ($P < 0.10$) & 144 (14.4\%) & 4 (0.4\%) \\
VTS reject ($P \geq 0.10$) & 826 (82.6\%) & 26 (2.6\%) \\
\bottomrule
\end{tabular}
\end{table}

The results reveal that VTS and dedup operate on almost entirely non-overlapping populations. Of the 852 ideas VTS would reject (probability $\geq 0.10$), 826 (96.9\%) are semantically novel; they would pass the dedup filter. This means VTS is not destroying useful diversity; it is filtering ideas that are predictable but distinct. Conversely, the dedup filter catches only 30 ideas (3.0\%), almost all of which VTS would also accept. The two mechanisms thus address complementary failure modes: VTS removes the ``obvious but novel'' ideas that inflate the concept space without adding depth, while dedup catches the rare near-duplicates that VTS misses.

\paragraph{VTS probability and typicality.} The strongest justification for VTS comes from the confusion matrix above: VTS and dedup target non-overlapping populations, with 96.9\% of VTS-rejected ideas being semantically novel. This population-level separation confirms that VTS removes a distinct class of candidates that dedup alone would miss. As a secondary check, we embed all 1{,}000 naive ideas with an independent model (\texttt{all-MiniLM-L6-v2}) and measure each idea's cosine distance to the distribution centroid. The Spearman correlation between self-assessed probability and centroid distance is significantly negative ($\rho = -0.065$, $p = 0.04$), indicating that higher self-assessed probability weakly predicts closer proximity to the distribution center. However, the signal is weak in absolute terms (Cohen's $d = 0.056$ between VTS-accepted and VTS-rejected groups; $R^2 < 0.005$). The practical value of VTS comes from the population-level separation shown in Table~\ref{tab:vts_confusion}, not from fine-grained probability calibration.

\subsection{Cross-Domain Validation}
\label{sec:cross_domain}

To test whether DCE generalizes beyond a single domain, we repeat the comparison on educational exam questions and creative writing prompts. Table~\ref{tab:cross_domain} summarizes results across all three domains.

\begin{table}[h]
\centering
\caption{Cross-domain results (200 batches each, seed 42; Table~\ref{tab:exp2} reports multi-seed means). DCE achieves 0\% collapse across all three domains. The educational domain exhibits the most severe naive collapse (34\%), making DCE's intervention most valuable there. Creative writing falls between the other two domains in natural redundancy.}
\label{tab:cross_domain}
\begin{tabular}{llccc}
\toprule
Domain & Method & Ideas accepted & EDV retention & Collapse rate \\
\midrule
Packaging & Naive & 1{,}000 & 27.0\% & 4.0\% \\
Packaging & VTS + dedup & 943 & 18.3\% & 0.0\% \\
Packaging & DCE & 950 & 28.5\% & 0.0\% \\
\midrule
Education & Naive & 1{,}000 & 16.8\% & 34.0\% \\
Education & VTS + dedup & 845 & 25.0\% & 0.0\% \\
Education & DCE & 919 & 26.2\% & 0.0\% \\
\midrule
Writing & Naive & 1{,}000 & 16.7\% & 8.4\% \\
Writing & VTS + dedup & 947 & 25.9\% & 0.0\% \\
Writing & DCE & 972 & 27.7\% & 0.0\% \\
\bottomrule
\end{tabular}
\end{table}

The three domains span a range of natural redundancy levels: packaging (4\% naive collapse), creative writing (8.4\%), and education (34\%). In all three, DCE achieves 0\% collapse and the highest EDV retention. The creative writing domain provides an intermediate test case: naive collapse is moderate but nontrivial, and DCE raises EDV retention from 16.7\% to 27.7\%, a 66\% relative improvement. The educational domain exhibits the worst naive collapse (34\%), suggesting that exam questions have a narrower natural concept distribution. Deduplication rejection rates are also higher across the board (15.5\% for VTS + dedup, 8.1\% for DCE), indicating that the model generates more repetitive content in this domain. Importantly, DCE still achieves 0\% collapse and the lowest rejection rate among dedup-enabled methods, confirming that adaptive prompting helps the model avoid duplicates proactively rather than relying solely on post-hoc rejection.

\subsection{Cross-Model Validation}
\label{sec:cross_model}

To test model generality, we run the packaging domain experiment with \texttt{claude-haiku-4-5} (Anthropic) using the same pipeline. Since Anthropic's API does not support OpenAI's structured output format, we provide the JSON schema in the system prompt and parse the response.

\begin{table}[h]
\centering
\caption{Model comparison on the packaging domain (200 batches). Claude Haiku has substantially higher dedup rejection rates than GPT-5-mini, indicating it generates more repetitive content. DCE's benefit is even more pronounced: it reduces rejection from 30.1\% to 11.0\%, a 19-point improvement.}
\label{tab:model_comparison}
\begin{tabular}{llcc}
\toprule
Model & Method & Ideas accepted & Dedup rejection rate \\
\midrule
GPT-5-mini & Naive & 1{,}000 & --- \\
GPT-5-mini & VTS + dedup & 943 & 5.7\% \\
GPT-5-mini & DCE & 950 & 5.0\% \\
\midrule
Claude Haiku 4.5 & Naive & 1{,}000 & --- \\
Claude Haiku 4.5 & VTS + dedup & 699 & 30.1\% \\
Claude Haiku 4.5 & DCE & 890 & 11.0\% \\
\bottomrule
\end{tabular}
\end{table}

Claude Haiku generates substantially more repetitive content than GPT-5-mini: VTS + dedup rejects 30.1\% of candidates (vs.\ 5.7\% for GPT-5-mini). This makes DCE's contribution even more valuable: adaptive prompting reduces rejection from 30.1\% to 11.0\%, a 19-percentage-point improvement. The model-agnostic nature of DCE's mechanisms (prompt construction, embedding-based dedup) means the pipeline transfers to new model families without modification.

\subsection{Sensitivity Analysis}
\label{sec:sensitivity}

We test sensitivity to all three key hyperparameters: the exploration-exploitation split, the VTS threshold $\tau$, and the dedup threshold $\delta$.

\paragraph{Exploration-exploitation split.} We vary the proportion of batches allocated to the exploration phase versus the exploitation phase, testing seven configurations from 0/100 (all exploitation) to 100/0 (all exploration).

\begin{table}[h]
\centering
\caption{Sensitivity to the exploration-exploitation split. All configurations achieve 0\% collapse, and EDV retention varies within a moderate range (22--30\%), indicating that DCE is robust to this parameter.}
\label{tab:sensitivity}
\begin{tabular}{lcc}
\toprule
Split (explore/exploit) & EDV retention & Collapse rate \\
\midrule
0/100 (all exploitation) & 22.2\% & 0.0\% \\
20/80 & 30.0\% & 0.0\% \\
30/70 & 23.2\% & 0.0\% \\
40/60 (default) & 27.1\% & 0.0\% \\
50/50 & 27.7\% & 0.0\% \\
60/40 & 26.9\% & 0.0\% \\
100/0 (all exploration) & 27.4\% & 0.0\% \\
\bottomrule
\end{tabular}
\end{table}

\paragraph{VTS threshold $\tau$.} We test $\tau \in \{0.05, 0.10, 0.20\}$, with $\tau = 0.10$ being the default.

\begin{table}[h]
\centering
\caption{Sensitivity to the VTS threshold $\tau$. All values achieve 0\% collapse. Lower $\tau$ is more aggressive (fewer ideas pass), while higher $\tau$ admits more ideas at the cost of slightly reduced depth.}
\label{tab:tau_sensitivity}
\begin{tabular}{lccc}
\toprule
$\tau$ & Ideas accepted & EDV retention & Collapse rate \\
\midrule
0.05 & 564 & 31.4\% & 0.0\% \\
0.10 (default) & 950 & 28.5\% & 0.0\% \\
0.20 & 941 & 27.0\% & 0.0\% \\
\bottomrule
\end{tabular}
\end{table}

The VTS threshold controls the aggressiveness of the probability filter. At $\tau = 0.05$, only 564 of 1{,}000 candidates survive, but these ideas achieve the highest EDV retention (31.4\%), confirming that filtering out ``obvious'' ideas genuinely improves the diversity of the accepted set. At $\tau = 0.20$, nearly all candidates pass (941), and EDV retention drops slightly to 27.0\%. All three settings achieve 0\% collapse, indicating that the dedup component alone is sufficient for collapse prevention regardless of VTS aggressiveness.

\paragraph{Dedup threshold $\delta$.} We test $\delta \in \{0.80, 0.85, 0.90, 0.95\}$, with $\delta = 0.85$ being the default.

\begin{table}[h]
\centering
\caption{Sensitivity to the dedup threshold $\delta$. Stricter thresholds (lower $\delta$) reject more ideas, improving diversity but reducing dataset size. Relaxed thresholds (higher $\delta$) retain more ideas while still preventing the worst collapse.}
\label{tab:delta_sensitivity}
\begin{tabular}{lccc}
\toprule
$\delta$ & Ideas accepted & EDV retention & Collapse rate \\
\midrule
0.80 & 869 & 27.5\% & 0.0\% \\
0.85 (default) & 950 & 28.5\% & 0.0\% \\
0.90 & 994 & 21.4\% & 0.0\% \\
0.95 & 999 & 25.5\% & 2.5\% \\
\bottomrule
\end{tabular}
\end{table}

\begin{figure}[h]
    \centering
    \IfFileExists{delta_tradeoff.pdf}{%
        \includegraphics[width=0.7\textwidth]{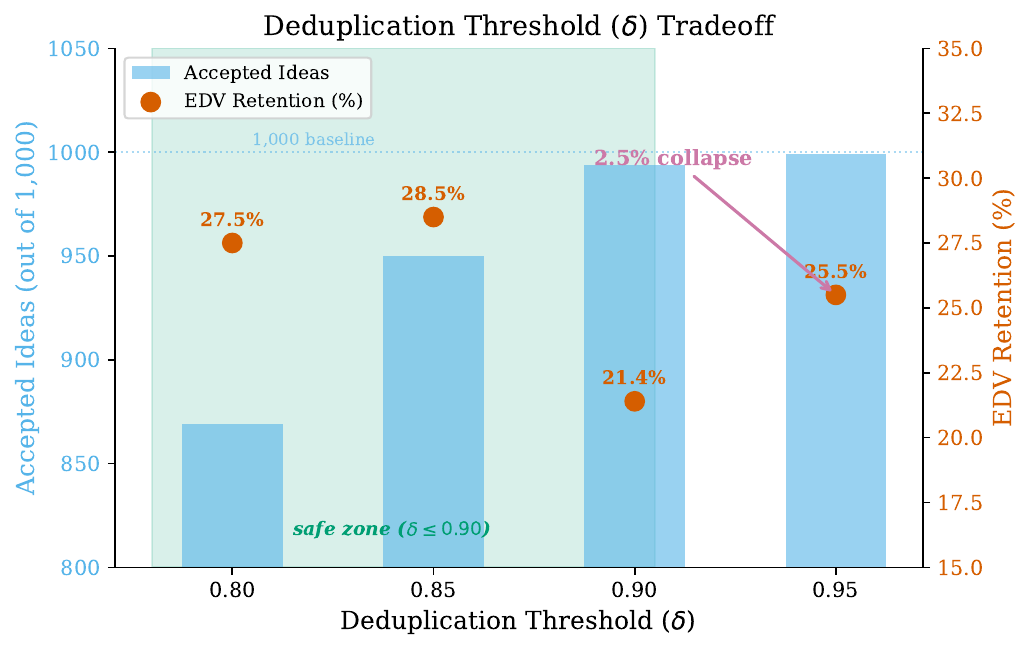}%
    }{%
        \fbox{\parbox{0.7\textwidth}{\centering\vspace{1.5cm}[Delta tradeoff figure: run generate\_delta\_tradeoff.py to generate]\vspace{1.5cm}}}%
    }
    \caption{Diversity-quantity tradeoff across $\delta$ values. Bars show accepted ideas (left axis); markers show EDV retention at each tested threshold (right axis). The dashed line marks the 1{,}000-idea baseline. At $\delta = 0.95$, collapse appears (2.5\%), marking the upper bound of safe relaxation.}
    \label{fig:delta_tradeoff}
\end{figure}

Figure~\ref{fig:delta_tradeoff} visualizes the diversity-quantity tradeoff across $\delta$ values.

The dedup threshold $\delta$ controls the tradeoff between dataset size and diversity. At $\delta = 0.80$, 131 ideas are rejected (869 accepted), maintaining 27.5\% EDV retention with 0\% collapse. At $\delta = 0.95$, nearly all ideas pass (999), but 2.5\% collapse appears (Figure~\ref{fig:delta_tradeoff}), confirming that $\delta = 0.95$ is too permissive for reliable collapse prevention. The default $\delta = 0.85$ achieves the best EDV retention (28.5\%) at 0\% collapse, but practitioners who need larger datasets can safely relax to $\delta = 0.90$ (994 ideas, 0\% collapse) at the cost of slightly reduced EDV retention.

\paragraph{Relaxed threshold downstream impact.} To test whether relaxed $\delta$ improves downstream task performance (where training set size matters), we train DeBERTa classifiers on DCE data generated at $\delta = 0.90$ across all three domains, with category count controlled at $K{=}5$.

The education domain provides the strongest evidence: at the default $\delta = 0.85$, DCE produces only 54 training examples (yielding 12.7\% F1 in the category-controlled evaluation). At $\delta = 0.90$, the training set grows to 164/41 train/val across 5 categories, and F1 jumps to 44.9\%, a 32-point improvement that confirms training set size, not diversity quality, was the binding constraint. In packaging, $\delta = 0.90$ produces 14.2\% F1 (328/82, 22 classes unconstrained), below the default 30.5\% (760/190, 9 classes), because relaxed dedup admits more categorically distinct ideas, spreading examples across 22 classes. The creative writing domain remains challenging: even at $\delta = 0.90$, only 52 training examples pass filtering (3.6\% F1), suggesting that this domain's fine-grained category structure requires either more batches or domain-specific tuning.

These results demonstrate that $\delta$ should be tuned per domain based on downstream requirements. High-redundancy domains like education benefit substantially from relaxed thresholds ($\delta = 0.90$), while low-redundancy domains like packaging perform best at the default $\delta = 0.85$.

\subsection{Per-Strategy Contribution}
\label{sec:per_strategy}

To evaluate the individual contribution of each diversity strategy, we analyze the 200 DCE batches (50 per strategy) and compute acceptance rate, mean EDV, and mean batch novelty for each.

\begin{table}[h]
\centering
\caption{Per-strategy performance in the packaging domain (seed 42, MiniLM embeddings). Constraint variation and cross-industry stimulus produce the highest novelty; gap targeting produces the lowest but serves a complementary role by steering toward underrepresented categories.}
\label{tab:per_strategy}
\begin{tabular}{lccc}
\toprule
Strategy & Accept \% & Mean EDV & Batch Novelty \\
\midrule
Gap targeting & 90.8\% & 0.265 & 0.277 \\
Assumption inversion & 94.0\% & 0.280 & 0.293 \\
Cross-industry stimulus & 97.2\% & 0.299 & 0.312 \\
Constraint variation & 98.0\% & 0.305 & 0.319 \\
\bottomrule
\end{tabular}
\end{table}

Constraint variation produces the highest acceptance rate (98.0\%) and EDV (0.305), while gap targeting has the lowest (90.8\%, 0.265). This is expected: gap targeting deliberately steers toward underpopulated regions where the model may struggle to produce novel ideas, while constraint variation and cross-industry stimulus provide the model with unusual framing that naturally produces more distinctive outputs. All four strategies contribute: the round-robin rotation ensures that gap targeting periodically rebalances the category distribution even as the other strategies drive novelty. Breaking results down by phase, exploration-phase batches outperform exploitation-phase batches across all strategies (mean EDV 0.330 vs.\ 0.259), confirming that the phase transition captures a genuine shift in generation difficulty as the concept space fills.

\subsection{Category Quality and Gap-Targeting Limitations}
\label{sec:category_quality}

To assess whether the model's self-assigned category labels are consistent enough to drive the gap-targeting strategy, we analyze the DCE packaging data. The model produces 707 unique category labels across 950 accepted ideas, with a normalized entropy of 0.978 (where 1.0 would indicate a perfectly uniform distribution), reflecting high label diversity with considerable fragmentation. Embedding-based analysis confirms that categories are nonetheless semantically coherent: the mean intra-category cosine similarity (0.648) substantially exceeds the mean inter-category centroid similarity (0.405), indicating that the model assigns similar labels to semantically similar ideas despite inconsistent surface forms.

\paragraph{Gap targeting with fragmented labels.} With 707 unique labels for 950 ideas (average count ${\sim}$1.34), nearly every category is ``underrepresented'' when gap targeting fires. The \texttt{get\_under\-represented\_\-categories()} function operates on raw model-generated labels with no coarse grouping, fuzzy matching, or aggregation. Gap targeting is most effective in the early campaign when the category space is sparse and categories are fewer and more meaningful. As the campaign progresses and label fragmentation increases, gap targeting's contribution diminishes, which is consistent with the per-strategy analysis showing gap targeting has the lowest acceptance rate and EDV (Table~\ref{tab:per_strategy}, 90.8\% accept, 0.265 EDV). The round-robin design ensures that the other three strategies, which do not depend on category labels, compensate. By batch 190, gap targeting is effectively selecting from a large pool of near-singleton categories, but it fires only every 4th batch, limiting the impact of this degradation.

\subsection{Diversity--Quantity Tradeoff Analysis}
\label{sec:downstream}

To assess how DCE's diversity filtering interacts with downstream utility, we train a DeBERTa-base~\citep{he2021deberta} text classifier on synthetic data from three sources: naive generation, VTS + dedup, and DCE. This evaluation probes the diversity--quantity tradeoff: aggressive filtering improves the conceptual diversity of accepted ideas but reduces training set size, and the net effect on downstream performance depends on which constraint binds first. The classification task uses coarse category labels extracted from the model-generated category field. Categories with fewer than 10 examples are dropped. Each model is trained for 3 epochs on an 80/20 stratified train/validation split and evaluated using weighted F1.

\paragraph{Controlled vs.\ unconstrained evaluation.} To partially isolate diversity quality from class structure effects, we present both a category-controlled comparison (fixed top-$K$ categories per method) and an unconstrained comparison (all categories) in Table~\ref{tab:downstream_combined}. The controlled evaluation holds class count constant, but training set sizes still vary across methods (e.g., packaging: 245/62 for naive vs.\ 152/38 for DCE; education: 355/89 vs.\ 54/14), so F1 differences reflect both data quality and sample size. The unconstrained evaluation reveals the full diversity-quantity interaction.

\begin{table}[h]
\centering
\caption{Downstream classification (DeBERTa-base, 3 epochs). Left columns: category-controlled (top-$K$); right columns: unconstrained. With class count held constant, DCE produces the best classifier in packaging and writing. In unconstrained mode, DCE doubles naive F1 on packaging (30.5\% vs.\ 15.2\%). Education is limited by training set size; relaxing $\delta$ to 0.90 resolves this (44.9\% F1).}
\label{tab:downstream_combined}
\begin{tabular}{ll cc cc}
\toprule
 & & \multicolumn{2}{c}{\textit{Controlled (top-$K$)}} & \multicolumn{2}{c}{\textit{Unconstrained}} \\
\cmidrule(lr){3-4} \cmidrule(lr){5-6}
Domain & Method & Train/Val & F1 & Train/Val & F1 \\
\midrule
Packaging & Naive & 245/62 & 17.1\% & 800/200 & 15.2\% \\
Packaging & VTS + dedup & 160/40 & 15.8\% & 754/189 & 6.2\% \\
Packaging & DCE & 152/38 & 19.3\% & 760/190 & 30.5\% \\
\midrule
Education & Naive & 355/89 & 83.8\% & 688/172 & 33.2\% \\
Education & VTS + dedup & 147/37 & 43.5\% & 508/127 & 7.2\% \\
Education & DCE & 54/14 & 12.7\% & 197/50 & 1.2\% \\
Education & DCE ($\delta{=}0.90$)$^\dagger$ & --- & --- & 164/41 & 44.9\% \\
\midrule
Writing & Naive & 356/89 & 13.2\% & 732/184 & 15.0\% \\
Writing & VTS + dedup & 181/46 & 10.8\% & 493/124 & 22.1\% \\
Writing & DCE & 90/23 & 14.2\% & 196/49 & 3.6\% \\
\bottomrule
\multicolumn{6}{l}{\scriptsize $^\dagger$Threshold-relaxation variant: dedup $\delta$ raised from 0.85 to 0.90 to increase training set size.}
\end{tabular}
\end{table}

With category count controlled (though not sample size), DCE produces the best classifier in two of three domains (packaging: 19.3\% vs.\ 17.1\% naive; writing: 14.2\% vs.\ 13.2\% naive), albeit with fewer training examples in both cases. In unconstrained mode, DCE doubles naive F1 on packaging (30.5\% vs.\ 15.2\%). The educational domain is limited by training set size: DCE at $\delta = 0.85$ produces only 197 training examples, but relaxing to $\delta = 0.90$ resolves this entirely (44.9\% F1), confirming that training set size, not diversity quality, was the binding constraint.

\paragraph{Takeaway.} DCE improves downstream classifier performance when the training set is large enough to support learning. In domains with high natural redundancy (education, 34\% collapse), relaxing $\delta$ to 0.90 recovers sufficient training examples while still preventing collapse (0\% at $\delta = 0.90$; Table~\ref{tab:delta_sensitivity}). Practitioners should tune $\delta$ per domain: use the default $\delta = 0.85$ for low-redundancy domains and consider $\delta = 0.90$ when downstream task performance is the primary objective.

\subsection{Embedding Space Visualization}
\label{sec:embedding_viz}

\begin{figure}[h]
    \centering
    \IfFileExists{embedding_space.pdf}{%
        \includegraphics[width=\textwidth]{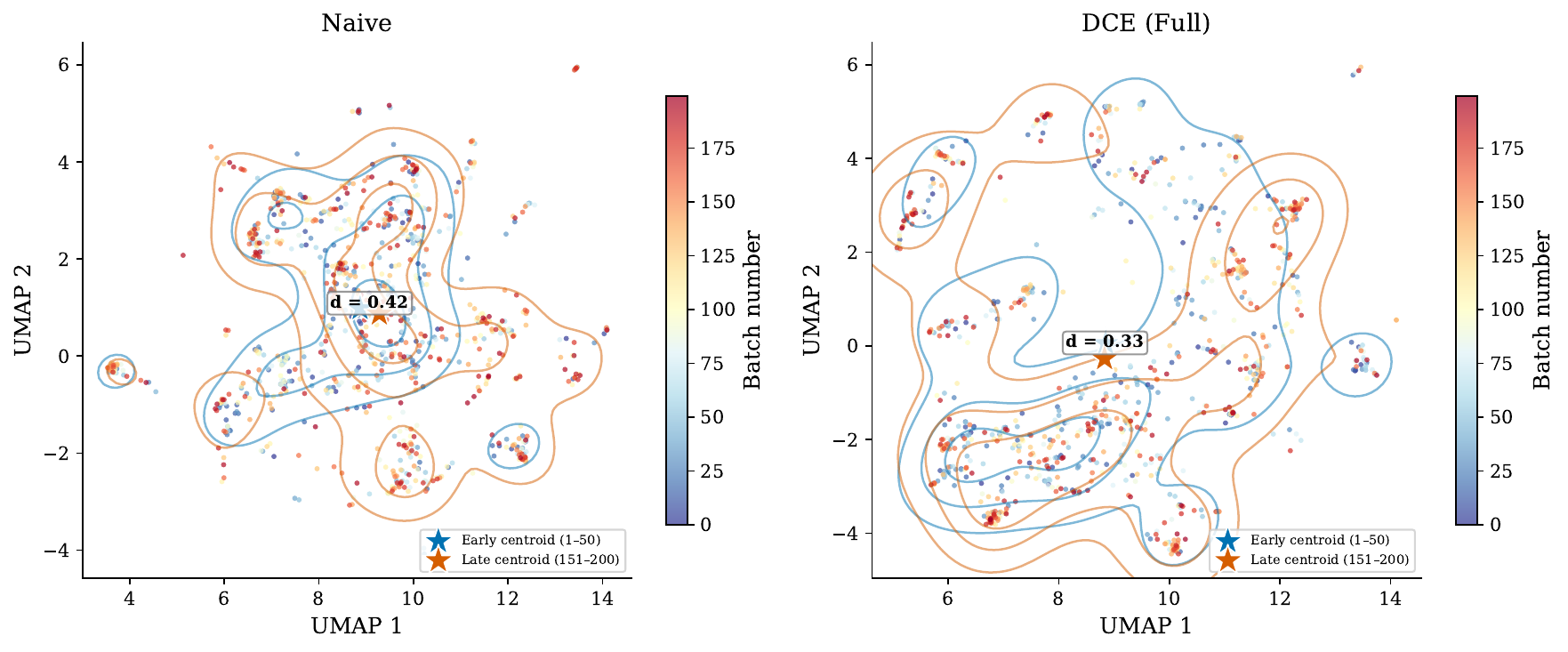}%
    }{%
        \fbox{\parbox{0.9\textwidth}{\centering\vspace{2cm}[UMAP embedding space visualization: run analyze\_embedding\_space.py to generate]\vspace{2cm}}}%
    }
    \caption{UMAP~\citep{mcinnes2018umap} projection of generated ideas under naive prompting (left) vs.\ DCE (right), colored by batch number on a diverging scale (blue = early, red = late). Density contours show the spatial extent of early-batch (blue, batches 1--50) and late-batch (red, batches 151--200) subsets. Under naive prompting, the contours overlap heavily, confirming mode collapse. Under DCE, the contours cover distinct territory, indicating sustained exploration. Star markers show subset centroids; $d$ is the Euclidean distance between them in UMAP space.}
    \label{fig:embedding_space}
\end{figure}

Figure~\ref{fig:embedding_space} provides a direct visual comparison of the embedding spaces produced by naive generation and DCE. Points are colored on a diverging scale (blue = early batches, red = late batches), with density contours showing the spatial extent of each subset. Under naive prompting, the early and late contours overlap heavily, confirming mode collapse. Under DCE, the contours cover distinct territory, demonstrating sustained conceptual exploration.

\subsection{Cost}

The full DCE pipeline generates 1{,}000 candidate ideas (5 per batch $\times$ 200 batches), of which 950 are accepted after VTS filtering and deduplication. The total API cost is approximately \$0.50--\$0.60, covering generation (\$0.50--\$0.59 across seeds for \texttt{gpt-5-mini} prompt and completion tokens) and embedding calls (under \$0.01 for \texttt{text-embedding-3-small}). Cost scales linearly with batch count; memory lookups and prompt construction introduce negligible overhead.

\section{Discussion}
\label{sec:discussion}

\paragraph{The interaction between filtering and steering.} The central finding of this work is that deduplication and prompt evolution are complementary because they address different failure modes. Adding deduplication to VTS without prompt evolution reduces EDV retention compared to VTS alone (Table~\ref{tab:exp2}, VTS + dedup at 18.3\% vs.\ VTS only at 23.8\%), because rejecting near-duplicates narrows the accepted set without redirecting the model toward new territory. Prompt evolution alone steers the model but cannot prevent semantic duplicates from accumulating (2.8\% collapse rate). Only when combined do they achieve sustained diversity with zero collapse.

\paragraph{VTS and dedup target different populations.} The confusion matrix analysis (Section~\ref{sec:vts_analysis}) reveals that VTS and dedup operate on largely non-overlapping sets of candidates. VTS removes high-probability ideas that may be semantically novel. Dedup removes near-duplicates regardless of their self-assessed probability. Removing either mechanism leaves a distinct failure mode unaddressed.

\paragraph{Cross-model and cross-domain generality.} DCE transfers to Claude Haiku 4.5 without modification, and the benefit is actually larger on this model (19-point reduction in rejection rate vs.\ 0.7 points for GPT-5-mini). The three-domain evaluation confirms that DCE eliminates collapse across domain structures with varying levels of natural redundancy.

\paragraph{When to deploy DCE versus simpler alternatives.} In low-redundancy domains where a few hundred ideas suffice, post-hoc deduplication may be sufficient: dedup-only achieves 0\% collapse at modest engineering cost. DCE is most valuable when: (a) the domain exhibits high natural redundancy (education at 34\% naive collapse, versus packaging at 4\%); (b) thousands of ideas are needed, where prompt evolution's steering prevents the model from exhausting its default repertoire; (c) conceptual breadth matters for the downstream task (DCE produces 18 HDBSCAN clusters vs.\ 9 for dedup-only); or (d) downstream classifier performance is the objective (DCE's 19.3\% F1 vs.\ 17.1\% for naive with controlled categories). Seed rotation with post-hoc deduplication (Section~\ref{sec:exp2}) provides a stronger baseline than naive generation alone, achieving 0\% collapse from 3$\times$ the candidate pool, but still produces lower EDV retention (21.6\% vs.\ DCE's multi-seed mean of 23.6\%) because it lacks the adaptive steering that drives DCE's diversity advantage. The cost-benefit calculation is straightforward: DCE adds approximately \$0.50 per 1{,}000 candidates and requires minimal engineering beyond the base generation pipeline.

\paragraph{Strategy degradation.} The per-strategy analysis (Section~\ref{sec:category_quality}) reveals that gap targeting degrades as the campaign progresses: with 707 unique labels for 950 ideas, nearly every category becomes a singleton, and gap targeting effectively selects from a large pool of equally ``underrepresented'' categories. This is reflected in the numbers: gap targeting achieves 90.8\% acceptance and 0.265 mean EDV, the lowest among the four strategies (Table~\ref{tab:per_strategy}). The round-robin design limits the impact, since gap targeting fires only every 4th batch and the other three strategies do not depend on category labels. However, an adaptive strategy selector that reduces gap targeting's frequency as label fragmentation grows, or a coarse-grouping step that clusters related labels before computing category gaps, could address this more directly. This connects to the fixed-rotation limitation noted in Section~\ref{sec:limitations}.

\paragraph{Token-level diversity is not concept-level diversity.} The impossibility of running temperature and nucleus sampling baselines with \texttt{gpt-5-mini} is itself informative. As model providers restrict decoding parameters, concept-level diversity interventions like DCE become the only available option. Even when token-level controls are available, they modify surface form variation without influencing which concepts the model selects.

\paragraph{Self-assessment as a practical filter.} The verbalized probability estimates used in VTS are not calibrated absolute probabilities. However, they are effective as relative rankings for separating obvious from non-obvious candidates, consistent with \citet{zhang2025verbalized}'s finding that model self-assessments preserve ordinal relationships despite poor absolute calibration.

\paragraph{Scalability considerations.} The current implementation performs a linear scan against all stored embeddings for deduplication. For the scale tested (1{,}000 accepted ideas), this is fast. At significantly larger scales (100{,}000+ ideas), approximate nearest neighbor indices would be required to maintain sub-linear lookup times.

\section{Limitations}
\label{sec:limitations}

\paragraph{Three domains.} We evaluate on sustainable packaging, educational exam questions, and creative writing prompts. While DCE eliminates collapse in all three, the downstream evaluation reveals domain-dependent behavior (Section~\ref{sec:downstream}). Structurally different tasks such as code generation, dialogue synthesis, or narrative writing remain untested.

\paragraph{Self-reported probability scores.} VTS relies on the model's own estimate of how predictable each idea is. These estimates are used as a relative filter, not a calibrated measure, but systematic biases in self-assessment could affect which ideas are retained.

\paragraph{API instability and reproducibility.} Results depend on specific model snapshots (\texttt{gpt\allowbreak-5\allowbreak-mini\allowbreak-2025\allowbreak-08\allowbreak-07}, \texttt{claude\allowbreak-haiku\allowbreak-4\allowbreak-5\allowbreak-20251001}). Providers may update or deprecate model versions without notice. Exact reproduction requires access to the same model versions. Full prompt templates and structured output schemas are provided in Appendix~\ref{sec:appendix_prompts}.

\paragraph{Fixed strategy rotation.} The four diversity strategies rotate on a rigid schedule. An adaptive approach that selects the strategy most likely to produce novel output given current memory state could yield further improvements.

\paragraph{Embedding space limitations.} Cosine similarity in a 1536-dimensional space is a lossy proxy for conceptual similarity. Two ideas may have similar embeddings but serve different functions, or vice versa. The $\delta = 0.85$ threshold is empirically chosen; our sensitivity analysis (Section~\ref{sec:sensitivity}) shows that results are robust across $\delta \in [0.80, 0.95]$, but different domains may benefit from different settings.

\paragraph{Diversity-quantity tradeoff.} As demonstrated in the educational domain, aggressive dedup filtering can reduce the accepted dataset to a size too small for effective downstream learning. Relaxing $\delta$ (Section~\ref{sec:sensitivity}) partially addresses this, but the optimal balance between diversity and quantity remains domain-dependent.

\paragraph{Limited seeds.} The main ablation (Table~\ref{tab:exp2}) uses 2--3 random seeds for the multi-seed methods (3 for naive, VTS + dedup, and DCE; 2 for VTS only). The collapse rate advantage (0.0 $\pm$ 0.0\% vs.\ 5.6 $\pm$ 2.0\%) and HDBSCAN cluster consistency (17--18 vs.\ 2--17) are the statistically robust findings; EDV retention differences ($p = 0.50$, Wilcoxon signed-rank) are directional but not significant at $n{=}3$. We accordingly treat collapse rate as the primary result, HDBSCAN clusters as the secondary measure of breadth, and EDV retention as supporting evidence. Cross-domain and sensitivity experiments use a single seed.

\section{Conclusion}
\label{sec:conclusion}

We have presented Dynamic Context Evolution, a framework for maintaining output diversity across large-scale synthetic data generation campaigns with language models. DCE addresses cross-batch mode collapse through three mechanisms: verbalized tail sampling filters predictable candidates, semantic memory prevents near-duplicate accumulation, and adaptive prompt evolution steers each batch toward unexplored conceptual territory.

Our experiments across three domains, two model families, and three random seeds demonstrate that these mechanisms are individually insufficient but jointly effective. The interaction between deduplication and prompt evolution is critical: filtering without steering narrows the output space, while steering without filtering allows semantic duplicates to accumulate. Together, they achieve 0.0 $\pm$ 0.0\% collapse (versus 5.6 $\pm$ 2.0\% for naive prompting) with the most consistent conceptual breadth (17--18 HDBSCAN clusters per seed versus naive's volatile 2--17). EDV retention is directionally higher for DCE but serves as supporting evidence given limited statistical power at $n{=}3$; the collapse rate and cluster structure results are the primary findings.

Sensitivity analysis over the VTS threshold $\tau$ and dedup threshold $\delta$ confirms that DCE is robust to parameter settings, while revealing that relaxed thresholds can address the diversity-quantity tradeoff in high-redundancy domains.

The framework generalizes across model families; the benefit is actually larger on Claude Haiku 4.5, which exhibits more repetitive behavior than GPT-5-mini. The UMAP visualization and HDBSCAN cluster analysis provide direct evidence that DCE's diversity improvements reflect genuine conceptual breadth, not merely metric artifacts.

The engineering requirements are modest: a vector database, a probability filter, and four prompt templates on a rotation schedule. No fine-tuning, reinforcement learning, or custom model architectures are required. The entire pipeline operates through standard API calls at a cost of approximately \$0.50 per 1{,}000 candidates.

\appendix
\section{Reproduction Details}
\label{sec:appendix_prompts}

This appendix provides the complete prompt templates and configuration required to reproduce our experiments. Code and experiment data are available at \url{https://github.com/ryanlingo/dynamic-context-evolution}.

\subsection{Hyperparameters}

\begin{table}[H]
\centering
\caption{Complete hyperparameter listing.}
\label{tab:hyperparams}
\begin{tabular}{llc}
\toprule
Component & Parameter & Value \\
\midrule
Generator (primary) & Model & \texttt{gpt-5-mini-2025-08-07} \\
Generator (primary) & Batch size & 5 \\
Generator (primary) & Temperature & (not configurable) \\
Generator (cross-model) & Model & \texttt{claude-haiku-4-5-20251001} \\
Generator (cross-model) & Max tokens & 4096 \\
Generator (cross-model) & Structured output & JSON schema in system prompt \\
Embeddings & Model & \texttt{text-embedding-3-small} \\
Embeddings & Dimensions & 1536 \\
VTS & Probability threshold ($\tau$) & 0.10 \\
Memory & Similarity threshold ($\delta$) & 0.85 \\
Memory & Recent ideas in prompt & 10 \\
Memory & Near-duplicates shown & 5 \\
Memory & Backend & ChromaDB \\
Prompt evolution & Strategy count & 4 \\
Prompt evolution & Phase threshold & 0.40 \\
HDBSCAN & min\_cluster\_size & 5 \\
HDBSCAN & metric & euclidean \\
Downstream & Model & \texttt{microsoft/deberta-base} \\
Downstream & Epochs & 3 \\
Downstream & Batch size & 16 \\
Downstream & Learning rate & $2 \times 10^{-5}$ \\
Downstream & Min examples per class & 10 \\
\bottomrule
\end{tabular}
\end{table}

\subsection{Prompt Templates}

Base generation prompt:
\begin{lstlisting}
You are {persona}. Generate exactly {batch_size} novel
{domain} ideas.

Each idea must be genuinely different from the others in
this batch and from the examples shown below.

{vts_instruction}
{strategy_instruction}
{phase_instruction}

## Recently accepted ideas (avoid repetition):
{recent_ideas}

## Near-duplicates to explicitly avoid:
{near_duplicates}

## Current category distribution:
{category_distribution}

Generate {batch_size} ideas now. For each idea, provide a
name, description, category, and your self-assessed
probability score.
\end{lstlisting}

VTS instruction (appended when VTS is active):
\begin{lstlisting}
IMPORTANT: For each idea, estimate the probability that
this specific idea would appear among all possible
responses to this prompt. Score it from 0.0 to 1.0.
We want ONLY unusual, surprising ideas -- aim for ideas
with probability below 0.10.
\end{lstlisting}

Strategy templates (one per batch, round-robin):
\begin{itemize}
    \item \emph{Gap targeting:} ``The following categories are underrepresented: \{categories\}. At least half of your ideas MUST target these.''
    \item \emph{Assumption inversion:} Lists recent idea assumptions; instructs model to invert each one.
    \item \emph{Cross-industry stimulus:} Samples 3 industries; asks ``What would \{domain\} look like if designed by someone from \{industry\}?''
    \item \emph{Constraint variation:} Applies a single extreme constraint (e.g., ``must work without electricity'').
\end{itemize}

\subsection{Structured Output Schema}

Each generation call uses OpenAI's structured output format (or equivalent JSON schema for Anthropic models) with the following schema:

\begin{lstlisting}
{
  "ideas": [
    {
      "name": "string",
      "description": "string (one paragraph)",
      "category": "string",
      "probability": "float (0.0-1.0)"
    }
  ]
}
\end{lstlisting}

\noindent enforced via Pydantic models with \texttt{beta.chat.completions.parse()} (OpenAI) or system prompt injection (Anthropic).

\bibliographystyle{plainnat}
\bibliography{references}

\begin{thebibliography}{16}
\providecommand{\natexlab}[1]{#1}
\providecommand{\url}[1]{\texttt{#1}}
\expandafter\ifx\csname urlstyle\endcsname\relax
  \providecommand{\doi}[1]{doi: #1}\else
  \providecommand{\doi}{doi: \begingroup \urlstyle{rm}\Url}\fi

\bibitem[Campello et~al.(2013)Campello, Moulavi, and
  Sander]{campello2013hdbscan}
Ricardo~JGB Campello, Davoud Moulavi, and J{\"o}rg Sander.
\newblock Density-based clustering based on hierarchical density estimates.
\newblock In \emph{Pacific-Asia Conference on Knowledge Discovery and Data
  Mining}, pages 160--172. Springer, 2013.

\bibitem[{Chroma}(2023)]{chromadb2023}
{Chroma}.
\newblock Chroma: The open-source embedding database.
\newblock \url{https://www.trychroma.com/}, 2023.

\bibitem[He et~al.(2021)He, Liu, Gao, and Chen]{he2021deberta}
Pengcheng He, Xiaodong Liu, Jianfeng Gao, and Weizhu Chen.
\newblock {DeBERTa}: Decoding-enhanced {BERT} with disentangled attention.
\newblock In \emph{International Conference on Learning Representations
  (ICLR)}, 2021.

\bibitem[Holtzman et~al.(2020)Holtzman, Buys, Du, Forbes, and
  Choi]{holtzman2020curious}
Ari Holtzman, Jan Buys, Li~Du, Maxwell Forbes, and Yejin Choi.
\newblock The curious case of neural text degeneration.
\newblock In \emph{International Conference on Learning Representations
  (ICLR)}, 2020.

\bibitem[Josifoski et~al.(2023)Josifoski, Sakota, Peyrard, and
  West]{josifoski2023exploiting}
Martin Josifoski, Marija Sakota, Maxime Peyrard, and Robert West.
\newblock Exploiting asymmetry for synthetic training data generation:
  {SynthIE} and the case of information extraction.
\newblock In \emph{Proceedings of the 2023 Conference on Empirical Methods in
  Natural Language Processing (EMNLP)}, pages 1555--1574, 2023.

\bibitem[Li et~al.(2024)]{li2024synthetic}
Haoran Li et~al.
\newblock Synthetic data (almost) from scratch: Generalized instruction tuning
  for language models.
\newblock \emph{arXiv preprint arXiv:2402.13064}, 2024.

\bibitem[McInnes et~al.(2018)McInnes, Healy, and Melville]{mcinnes2018umap}
Leland McInnes, John Healy, and James Melville.
\newblock {UMAP}: Uniform manifold approximation and projection for dimension
  reduction.
\newblock \emph{arXiv preprint arXiv:1802.03426}, 2018.

\bibitem[Neelakantan et~al.(2022)Neelakantan, Xu, Puri, Radford, Han, Tworek,
  Yuan, Tezak, Kim, Hallacy, et~al.]{neelakantan2022text}
Arvind Neelakantan, Tao Xu, Raul Puri, Alec Radford, Jesse~Michael Han, Jerry
  Tworek, Qiming Yuan, Nikolas Tezak, Jong~Wook Kim, Chris Hallacy, et~al.
\newblock Text and code embeddings by contrastive pre-training, 2022.

\bibitem[Reimers and Gurevych(2019)]{reimers2019sentencebert}
Nils Reimers and Iryna Gurevych.
\newblock Sentence-{BERT}: Sentence embeddings using siamese {BERT}-networks.
\newblock In \emph{Proceedings of the 2019 Conference on Empirical Methods in
  Natural Language Processing (EMNLP)}, pages 3982--3992, 2019.

\bibitem[Settles(2009)]{settles2009active}
Burr Settles.
\newblock \emph{Active Learning Literature Survey}.
\newblock University of Wisconsin-Madison Department of Computer Sciences,
  2009.

\bibitem[Shumailov et~al.(2024)Shumailov, Shumaylov, Zhao, Papernot, Anderson,
  and Gal]{shumailov2024ai}
Ilia Shumailov, Zakhar Shumaylov, Yiren Zhao, Nicolas Papernot, Ross Anderson,
  and Yarin Gal.
\newblock Ai models collapse when trained on recursively generated data.
\newblock \emph{Nature}, 631:\penalty0 755--759, 2024.

\bibitem[Snoek et~al.(2012)Snoek, Larochelle, and Adams]{snoek2012practical}
Jasper Snoek, Hugo Larochelle, and Ryan~P Adams.
\newblock Practical {B}ayesian optimization of machine learning algorithms.
\newblock In \emph{Advances in Neural Information Processing Systems},
  volume~25, 2012.

\bibitem[Wang et~al.(2023)Wang, Wei, Schuurmans, Le, Chi, Narang, Chowdhery,
  and Zhou]{wang2023selfconsistency}
Xuezhi Wang, Jason Wei, Dale Schuurmans, Quoc Le, Ed~Chi, Sharan Narang,
  Aakanksha Chowdhery, and Denny Zhou.
\newblock Self-consistency improves chain of thought reasoning in language
  models.
\newblock In \emph{International Conference on Learning Representations
  (ICLR)}, 2023.

\bibitem[Wu et~al.(2023)Wu, Zhu, Albayrak, Axon, Bertsch, Deng, Ding, Guo,
  Gururaja, Kuo, et~al.]{long2024llms}
Tongshuang Wu, Haiyi Zhu, Maya Albayrak, Alexis Axon, Amanda Bertsch, Wenxing
  Deng, Ziqi Ding, Bill Guo, Sireesh Gururaja, Tzu-Sheng Kuo, et~al.
\newblock {LLMs} as workers in human-computational algorithms? replicating
  crowdsourcing pipelines with {LLMs}.
\newblock \emph{arXiv preprint arXiv:2307.10168}, 2023.

\bibitem[Yu et~al.(2023)Yu, Zhuang, Zhang, Meng, Ratner, Krishna, Shen, and
  Zhang]{yu2024large}
Yue Yu, Yuchen Zhuang, Jieyu Zhang, Yu~Meng, Alexander Ratner, Ranjay Krishna,
  Jiaming Shen, and Chao Zhang.
\newblock Large language model as attributed training data generator: A tale of
  diversity and bias.
\newblock In \emph{Advances in Neural Information Processing Systems},
  volume~36, 2023.

\bibitem[Zhang et~al.(2025)Zhang, Yu, Chong, Sicilia, Tomz, Manning, and
  Shi]{zhang2025verbalized}
Jiayi Zhang, Simon Yu, Derek Chong, Anthony Sicilia, Michael~R. Tomz,
  Christopher~D. Manning, and Weiyan Shi.
\newblock Verbalized sampling: How to mitigate mode collapse and unlock {LLM}
  diversity.
\newblock \emph{arXiv preprint arXiv:2510.01171}, 2025.
\newblock URL \url{https://www.verbalized-sampling.com/}.

\end{thebibliography}

\end{document}